\documentclass[lettersize,journal]{IEEEtran}
\usepackage{amsmath,amsfonts}
\usepackage{algorithmic}
\usepackage{algorithm}
\usepackage{array}
\usepackage[caption=false,font=normalsize,labelfont=sf,textfont=sf]{subfig}
\usepackage{textcomp}
\usepackage{stfloats}
\usepackage{url}
\usepackage{verbatim}
\usepackage{graphicx}
\usepackage{cite}
\usepackage{color}
\usepackage{hyperref}
\hypersetup{colorlinks=true, linkcolor=blue}
\usepackage{times}
\usepackage{epsfig}
\usepackage{graphicx}
\usepackage{amsmath}
\usepackage{amssymb}
\usepackage{booktabs}
\usepackage{multirow}
\definecolor{myPurple}{rgb}{0.4, .0, .8}
\definecolor{myGreen}{rgb}{0, .8, .3}
\definecolor{myRed}{rgb}{0.8, .2, .2}
\definecolor{myPink}{rgb}{0.8, .5, .5}
\definecolor{myOrange}{rgb}{0.7, 0.45, 0.2}
\definecolor{myBlue}{rgb}{.0, .0, 1.0}
\definecolor{myBlue2}{rgb}{.0, .0, 0.5}
\definecolor{myBlack}{rgb}{.0, .0, 0.0}
\definecolor{mycyan}{rgb}{.39,.58,.93}

\begin{document}

\title{CodingHomo: Bootstrapping Deep Homography with Video Coding}

\author{
        ~Yike~Liu,~\IEEEmembership{Student Member,~IEEE,}
        ~Haipeng~Li,~\IEEEmembership{Student Member,~IEEE,}
        ~Shuaicheng~Liu, ~\IEEEmembership{Senior Member,~IEEE,}
        and Bing~Zeng,~\IEEEmembership{Fellow,~IEEE}

\IEEEcompsocitemizethanks{

\IEEEcompsocthanksitem Manuscript submitted on February 5, 2024; revised April 23, 2024 and May 27, 2024; accepted June 16, 2024. This work was supported in part by the National Natural Science Foundation of China (NSFC) under Grant No. 62372091 and No. 62031009, in part by the Sichuan Science and Technology Program of China under grant No.2023NSFSC0462.
\IEEEcompsocthanksitem Authors are with School of Information and Communication Engineering, University of Electronic Science and Technology of China, Chengdu, Sichuan, China.
\IEEEcompsocthanksitem Corresponding authors: Shuaicheng Liu (liushuaicheng@uestc.edu.cn) and Bing Zeng (eezeng@uestc.edu.cn)


}}



\maketitle

\begin{abstract}
Homography estimation is a fundamental task in computer vision with applications in diverse fields. Recent advances in deep learning have improved homography estimation, particularly with unsupervised learning approaches, offering increased robustness and generalizability. However, accurately predicting homography, especially in complex motions, remains a challenge. In response, this work introduces a novel method leveraging video coding, particularly by harnessing inherent motion vectors (MVs) present in videos. We present CodingHomo, an unsupervised framework for homography estimation. Our framework features a Mask-Guided Fusion (MGF) module that identifies and utilizes beneficial features among the MVs, thereby enhancing the accuracy of homography prediction. Additionally, the Mask-Guided Homography Estimation (MGHE) module is presented for eliminating undesired features in the coarse-to-fine homography refinement process. CodingHomo outperforms existing state-of-the-art unsupervised methods, delivering good robustness and generalizability. The code and dataset are available at: \href{github}{https://github.com/liuyike422/CodingHomo}
\end{abstract}


\section{Introduction}

A homography is a $3\times3$ transformation matrix that defines the perspective relationships of a planar surface between two images~\cite{dlt}. Estimating homography is an essential task in computer vision with a wide array of applications, including but not limited to, autonomous driving~\cite{gu2022homography, wang2023homography, fan2021learning}, medical imaging~\cite{li2021homography, huber2021homography, allan2021stereo}, image restoration~\cite{HDR1, HDR2, HDR3, liu2014fast, bhat2021deep}, video stabilization~\cite{stabilization1, stabilization2}, video captioning~\cite{yan2022gl, yan2022video}, video segmentation~\cite{yan2022solve, liu2021sg} and video object detection~\cite{cui2021tf, liu2020video}. Traditional methods extract image feature matches~\cite{sift, surf, orb} and then solve the direct linear transformation (DLT)~\cite{dlt} with outlier rejection, e.g., RANSAC~\cite{ransac}, for the homography estimation. However, the performance heavily depends on the quality of the extracted key points and the precision of the matching correspondences. They lose robustness in environments with challenging conditions (such as low-light, adverse weather or presence of dynamic objects)~\cite{cahomo, gyroflow+}, leading to performance degradation in the downstream applications.

\begin{figure}[t]
\begin{center}
    \includegraphics[width=1\linewidth]{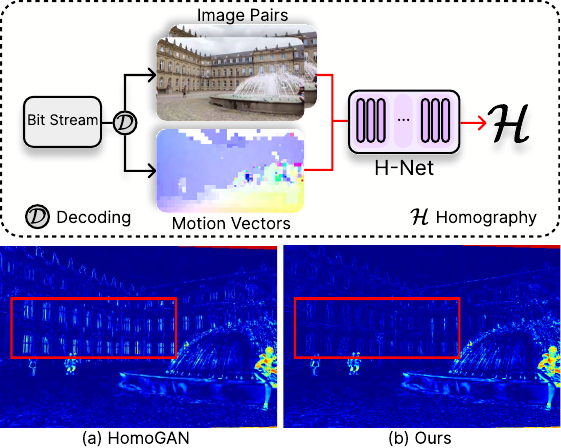}
\end{center}
  \caption{The overview of our work. We extract MVs while decoding frames and utilize them along with reconstruct images as input for homography estimation. We illustrate the error heatmap between target and warped source image in the (a) and (b), the darker the image, the better the alignment. Our result significantly outperforms existing method~\cite{homogan} in dynamic foregrounds scene with the prior MVs.}
  \label{fig_teaser}
\end{figure}

Recent advancements in deep learning-based methods have improved the performance of homography estimation, even under adverse conditions. These methods can be classified into two main categories: supervised and unsupervised learning approaches. Supervised learning methods~\cite{dhn, localtrans, ihn, realsh} have been trained using pairs generated by augmenting single images, such as those in MS-COCO dataset~\cite{lin2014microsoft}. Although these methods are effective in certain scenarios, they fall short in generalizability due to the domain gap between the augmented training datasets and real-world situations. Conversely, unsupervised learning methods~\cite{nguyen2018unsupervised, cahomo, baseshomo, homogan} directly minimize the photometric loss in the deep feature space between the warped source and target images. Benefiting from feature losses~\cite{cahomo, baseshomo} and masks~\cite{homogan} that reject outliers, these methods have shown overall improvement in both regular and adverse conditions. However, accurately predicting homography, especially in complex motions, still remains a substantial challenge. In this work, we address this challenge by leveraging the motion prior in video coding.


Video coding is designed to maximize the compression of video data while preserving as much original information as possible. Due to the substantial size, most videos undergo compression before they are stored or distributed. It is important to note that motion vectors (MVs) play a critical role in achieving video compression as tracking the movement of pixel blocks from a reference frame to the current frame, facilitating effective data reduction. During the decoding process, these MVs are used to warp the reference frame, which, along with the residual data, allows for the accurate reconstruction of target frames. While the primary purpose of MVs in video coding is to maximizing compression performance, research has shown that they can also be instrumental in estimating motion~\cite{codingflow}. Building on the two main reasons: 1) MVs have the potential to enhance global motion estimation, and 2) Since MVs are naturally embedded in coded videos and can be extracted with no additional overhead, we propose utilizing MVs for homography estimation.

\begin{figure}[t]
\begin{center}
    \includegraphics[width=1\linewidth]{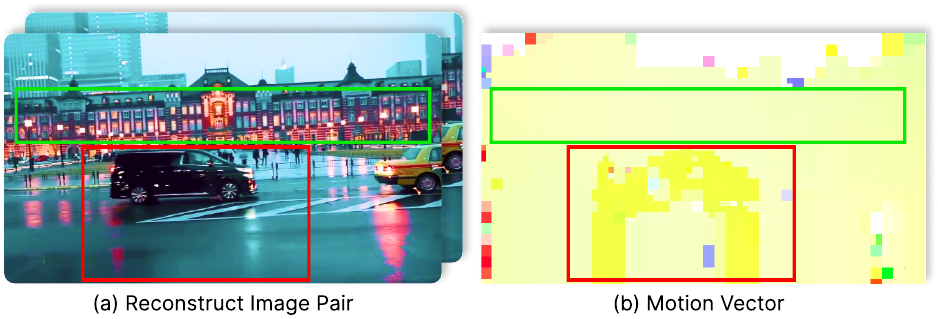}
\end{center}
  \caption{An example of network input. (a) Reconstruct image pair. (b) MVs. Green block indicates the dominant plane area. Red block donates a dynamic vehicle. }
  \label{fig_vis_input}
\end{figure}

In this paper, we introduce CodingHomo, an innovative unsupervised framework for homography estimation that utilizes MVs obtained from video encoding. Our approach begins with the extraction of inter MVs from the coded bitstream, alongside the reconstruction of images. These images, together with the MVs, serve as input to our network, aiming at predicting homography as indicated in Fig.~\ref{fig_teaser}. We find that CodingHomo improves homography estimation performance in complex conditions compared to the previous state-of-the-art (SOTA) unsupervised method, exemplified in the second row of Fig.\ref{fig_teaser}. Here, we utilize a heatmap to visualize alignment results, where darker colors indicate better alignment. Diving into the technical core, we develop a Mask-Guided Fusion (MGF) module that incorporates MVs into the homography estimation process. This module not only extracts advantageous features (as highlighted in green box at Fig.~\ref{fig_vis_input}) but also identifies and filters out outliers (as indicated in red box at Fig.~\ref{fig_vis_input}), using a motion reject mask. Furthermore, we refine the homography estimation process with a Mask-Guided Homography Estimation (MGHE) module, enhancing the coarse-to-fine mechanism by applying MGF-derived masks to remove undesirable image features. Additionally, the introduction of the Enhanced Motion Mask (EMM) strategy employs a probabilistic method, combined with image feature masks, to refine loss calculation and thus improve framework's accuracy. Overall, our primary contributions include:

\begin{itemize}
    \item We present the first unsupervised deep homography framework based on MVs extracted from video coding. 
    \item The development of the MGF and MGHE modules for effective extraction and integration of viable motion information from MVs into the homography estimation process. Moreover, the EMM strategy considerably focuses the loss function on improving homography prediction accuracy.
    \item We complement large-scale homography datasets with MVs, and experimental results demonstrate that our framework achieves state-of-the-art performance, delivering superior robustness generalizability.
\end{itemize}

\section{Related Work}

\subsection{Video Coding}

Video coding technology compresses video by eliminating redundancy in raw video. The main categories of redundancies are temporal, spatial, symbolic, and visual. Traditional video coding technology~\cite{h264, hevc, vvc} employs a block-based hybrid coding architecture that includes the following processes: prediction, transformation, quantization, and entropy coding. Prediction technology, which reduces spatial and temporal redundancies, has two main modes: intra-frame and inter-frame prediction. The transformation process, utilizing techniques such as the Discrete Cosine Transform (DCT)~\cite{dct}, plays a important role by converting the residual block after prediction from the pixel domain to the frequency domain, thereby concentrating most of the block's energy into a few significant coefficients. Quantization, which is crucial for eliminating visual redundancy, involves a delicate balance between compression ratio and reconstructed image quality. The varying quantization levels not only affect compression efficacy but also influence the video recover quality. To minimize symbolic redundancy, entropy coding such as Context Adaptive Binary Arithmetic Coding (CABAC)~\cite{cabac} encodes frequently occurring patterns with fewer bits.
Additionally, Loop Filters are incorporated into the coding framework to enhance the quality of the decoded video by mitigating artifacts such as blocking~\cite{deblocking}. Recently, several deep learning-based video compression algorithms~\cite{dvc, m-lvc, fvc} have emerged, aiming to deliver enhanced compression performance. Nevertheless, their practical application is currently constrained by substantial computational demands and limited generalization across diverse hardware platforms. 

Meanwhile, Motion-Compensated Prediction (MCP)~\cite{mcp1, mcp2} is a key component of the inter coding mode. In MCP, MVs denote the positional relationship between the current block and the prediction block in reference frames. The process of determining the optimal MV is termed Motion Estimation (ME). While an exhaustive search based on the mean square error criterion can precisely identify the best MV, it comes at the expense of significant computational overhead. To address this issue, a variety of fast search algorithms—such as the Three-Step Search (TSS)~\cite{tss}, the New Three-Step Search (NTSS)~\cite{ntss}, the Four-Step Search (4SS)~\cite{fss}, and the Diamond-Shaped Search~\cite{dia_s} have been developed and are commonly implemented in video codecs to reduce complexity.

\begin{figure*}[!ht]
\begin{center}
    \includegraphics[width=0.95\linewidth]{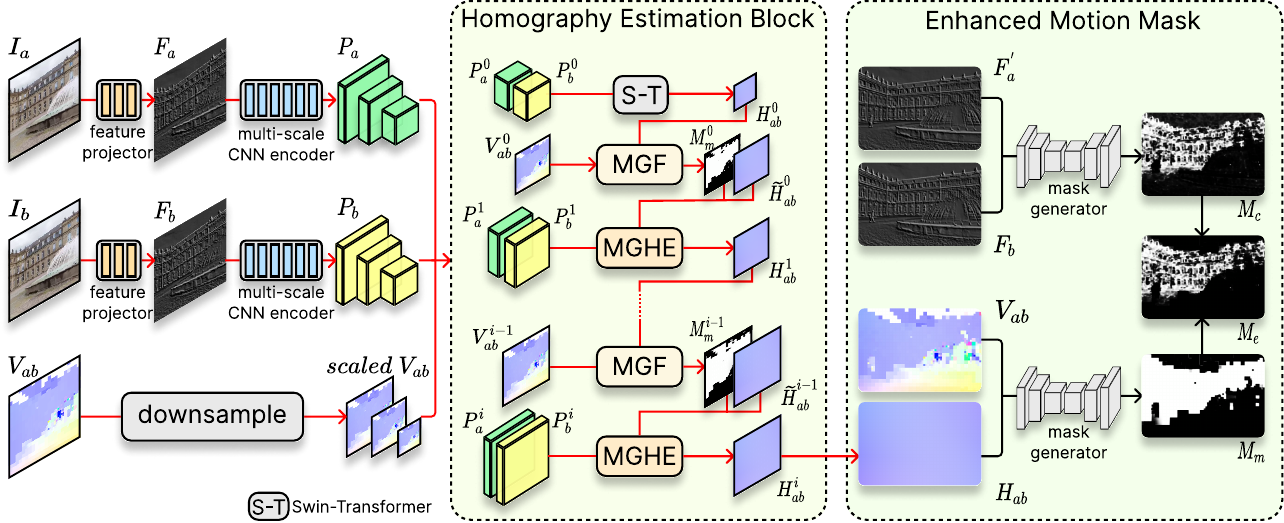}
\end{center}
  \caption{The overall pipeline of CodingHomo. Our network architecture consists of three stages: 1) Feature extraction stage. A CNN module for projecting input images into feature space and a multi-scale CNN encoder for generating feature pyramid. 2) Homography estimation stage. A block with cascaded MGF and MGHE blocks for predicting the homography from coarse to fine. 3) Mask prediction stage. A mask generated by $V_{ab}$, $H_{ab}$, $F_b$ and warped $F_a$($F'_a$) is applied to loss function to help the network focusing on dominant plane. Red arrows indicate the inference pipeline.}
  \label{fig_pip}
\end{figure*}

\subsection{Homography Estimation}

Conventional homography estimation methods typically involve three stages: feature detection, using algorithms such as SIFT~\cite{sift} or ORB~\cite{orb}; correspondence matching~\cite{cunningham2021k}; and outlier rejection techniques like RANSAC~\cite{ransac}. Recent advances in learning-based feature detection and matching algorithms, including LIFT~\cite{lift}, SuperPoint~\cite{superpoint}, and SOSNet~\cite{sosnet}, have improved the robustness and effectiveness of homography estimation. Furthermore, outlier rejection techniques such as MAGSAC~\cite{magsac} and MAGSAC++~\cite{barath2020magsac++} have made algorithms more reliable, even in complex scenarios that involve multiple planes, parallax, and dynamic foregrounds.

Optimization-based methodologies~\cite{Clkn},~\cite{DPCP}, which utilize strategies like Lucas-Kanade or direct computation of sum of squared differences, estimate homography by iteratively refining a randomly initialized homography. Nevertheless, these approaches can be time-consuming and prone to cumulative errors. The pursuit of deep image alignment techniques, such as optical flow~\cite{gyroflow,gyroflow+,ASFlow,realflow,KPFlow} and dense correspondence~\cite{pdc, GLU-Net}, led to the first deep homography estimation network, proposed through supervised learning in 2016~\cite{dhn}.

Deep learning-based methods for homography estimation fall into two categories: supervised and unsupervised. Supervised approaches~\cite{Dynamic-supervised2020, localtrans, ihn, jiang2023semi} use synthetic image pairs extracted from single images for training but are disadvantaged by a lack of dynamics and realistic scene parallax, which limits their capacity to generalize to real-world conditions. Recently RealSH~\cite{realsh} address the issue by generating a realistic dataset. In contrast, unsupervised methods~\cite{nguyen2018unsupervised, cahomo, baseshomo, homogan} exhibit increased robustness through their label-free training regimes. Nonetheless, they tend to struggle with dynamic objects and homogeneous regions~\cite{GLU-Net} due to an over-reliance on photometric loss. The CAHomo method~\cite{cahomo} addresses this by introducing a self-guided mask to accentuate key-point features. BasesHomo~\cite{baseshomo} improves homography estimation by constraining the rank of feature maps and applying a feature-identity loss, using an 8-weighted basis set to implicitly bolster results. HomoGAN~\cite{homogan} uses Generative Adversarial Network (GAN) loss to isolate the dominant plane and applies a Transformer encoder for a coarse-to-fine strategy.

In this work, we bootstrap deep homography methods with the help of MVs from video coding.

\subsection{Deep Video Vision Tasks}

Recent studies have made significant advancements in video vision tasks using deep learning. Yan et al.~\cite{yan2022video} devised the Global-Local Representation Granularity (GLR) framework with a global-local encoder and a progressive training strategy, achieving competitive video captioning performance. Liu et al.~\cite{liu2021sg} proposed a one-stage spatial granularity network (SGNet) for video instance segmentation, featuring a compact architecture, interdependent task heads, dynamic mask prediction, avoidance of expensive proposal-based features, and a robust tracking head, resulting in state-of-the-art performance and inference speed. Cui et al.~\cite{cui2021tf} introduced TF-Blender for video object detection, enhancing per-frame representation by modeling lower-level temporal relations and combining enriched feature maps, demonstrating improved performance. Additionally, recent studies have explored the utility of MVs in various vision tasks. Chen et al.~\cite{super-resolution} improved compressed video super-resolution by integrating decoding prior and deep prior interactions. Zhang et al.~\cite{actionrecognition} developed a real-time action recognition system based on MVs instead of optical flow, achieving notable speed enhancements. Zhu et al.~\cite{zhu2024cpga} used coding information, including MVs, as priors for video quality enhancement, achieving superior outcomes. Tan et al.~\cite{segmentation} investigated a propagation-based segmentation technique in the compressed domain to boost inference speed, while Zhou et al.~\cite{mvflow} utilized MVs to enhance optical flow estimation performance. Our distinct focus lies in homography, emphasizing motion within the dominant plane.

\section{Method}

\subsection{Overview}

In this section, we introduce a novel framework, CodingHomo, that combines motion vector $V_{ab}$ extracted from coding bit stream and corresponding images $I_a$ and $I_b$ to jointly estimate an homography transformation $H_{ab}$. The overall pipeline of our method is illustrated in Fig.~\ref{fig_pip}. 

First, the feature extraction stage consists of two elements: the first one is the feature projector $\mathcal{P}$~\cite{cahomo} which is leveraged to transform the input images $I_a$ and $I_b$ into single-channel feature maps $F_a$, $F_b$ $ \in \mathbf{R}^{1\times H \times W}$ for feature loss computation; the second one is the multi-scale CNN encoder~\cite{homogan} designed for producing feature pyramids to achieve coarse-to-fine homography estimation. Notably, motion vector $V_{ab}$ is respectively scaled to fit the size of feature pyramids. 

In the second stage, we introduce a homography estimation block comprising two key modules: a mask-guided fusion module (MGF) (Sec. \ref{MGF}) and a mask-guided homography module (MGH) (Sec. \ref{MGHE}), both specifically tailored for accurate homography estimation. The MGF and MGH blocks are cascaded to predicts the homography from coarse to fine.

Finally, we propose a novel dominant plane mask $M_{e}$ (Sec. \ref{Mcmc}), generated by $V_{ab}$, $H_{ab}$, $F_{a}$ and $F_{b}$. This mask is incorporated into our total loss function. We optimize the entire model by minimizing an unsupervised objective function, which allows us to achieve a robust and reliable homography estimation.

\begin{figure}[t]
\begin{center}
  \includegraphics[width=1\linewidth]{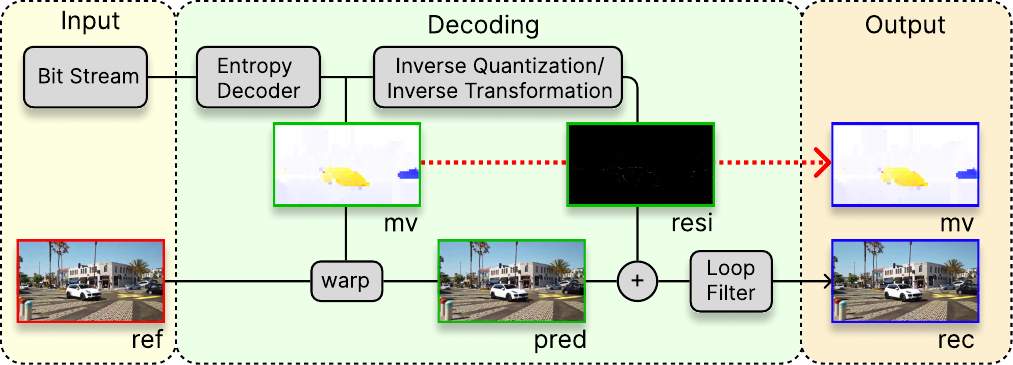}
\end{center}
  \caption{Illustration of the extraction of MVs during the decoding process. The input includes a reference frame and the bit stream of the current frame. Initially, the intermediate data of decoding process such as  residual data and MVs, are typically discarded after decoding. In our approach, we preserve the MVs and output them along with the reconstructed frame. 
  }
  \label{fig_motion_extract}
\end{figure}

\subsection{Motion Vector Extraction}

In this section, we will discuss the extraction of motion vectors (MVs) from encoded videos and examine reasons why certain MVs might introduce undesired interference when utilized as alignment priors.

Although newly published video video coding standards such as HEVC/H.265~\cite{hevc} and VVC/H.266~\cite{vvc} offer enhanced performance in video compression, the AVC/H.264~\cite{h264} standard remains widely utilized, particularly in hardware devices, due to its lower complexity. Our work is based on the AVC/H.264 standard because the CA-unsup dataset~\cite{cahomo} collects image pairs from real-world video frames encoded in this format. This methodology has the potential to be modified for use with the latest compression standards. In the AVC/H.264 scheme, each frame is divided into macroblocks of size 16x16. While macroblocks can be encoded using either intra or inter mode, our research focuses on the inter mode, which reveals the relationship between the current block and a reference block.

In the AVC/H.264 codec, a tree-structured strategy enables macroblocks of size 16x16 to be further subdivided into smaller blocks, such as 16x8, 8x16, 8x8, 8x4, 4x8, and 4x4. However, due to the computational complexity of the process, some hardware encoders limit the minimum partition size to 8x8, as is the case with the CA-unsup dataset. During the encoding phase, each candidate partition undergoes several stages including motion estimation, motion compensation, transformation, quantization, and finally, entropy coding is performed to generate the bit stream. To obtain the reconstructed result necessary for distortion calculations, inverse quantization and transformation are applied. Finally, each partition's Rate-Distortion (RD) cost is calculated based on:
\begin{equation}
\label{RDO}
RDcost = distortion  + \lambda * rate.
\end{equation}
The RD cost comprises the bit rate, which is the number of bits used, and distortion, which is the mean square error between the original and reconstructed blocks. The parameter $\lambda$ acts as a weighting factor to balance the two. The partition yielding the lowest RD cost is selected for use in the encoded video.

It is worth noting that the optimal match between a current block and a reference block typically results in the least total residual energy, which tends to require fewer bits following transformation, quantization, and entropy coding. However, the quantity of bits necessary to encode the motion vector information can influence the overall encoding strategy. Consequently, the motion vector selected during the final encoding process may not always correspond to the one that would yield the optimal image match. For example, MVs from a merge list (which use the same MVs as previously coded neighboring blocks) or zero-value MVs (indicating the same position in both current and reference blocks) are often favored due to their low bit costs. Despite possibly higher residual energy, these strategies can result in lower RD costs overall.

In the decoding process, we extract all MVs while reconstructing the image, as Fig.~\ref{fig_motion_extract} illustrates the simplicity of this extraction. We consider both the previously decoded reference frame and the current frame's bit stream as input sources. For inter-prediction blocks, information such as the motion vector is entropy-decoded from the bit stream and utilized for motion compensation to construct the predicted block. The residual is generated from the bit stream through processes that include entropy decoding, inverse quantization, and inverse transformation, and is then added to the prediction block. Subsequently, a loop filter is applied to enhance image quality. The decoding process can be described as: 
\begin{equation}
\label{decode}
rec = \mathcal{LF}(resi+\mathcal{W}(ref, MVs)),
\end{equation}
where $rec$ is the reconstruction image,  $\mathcal{LF}$ is loop filtering operation, $resi$ is the residual, $\mathcal{W}$ is the warping operation and $ref$ is the reference frame. Intermediate result normal be deleted at the end of reconstruction, we extract MVs from the template buffer. The extracted data is classified into three types: the coordinates of the block within the current frame, the motion vector itself, and the reference frame index. The storage unit of MVs is sized as an 8x8 block, matching the minimum inter prediction unit size in the CA-unsup dataset. For inter blocks with larger size, their MVs are replicated across all individual 8x8 units within the block. Before sending to the network, MVs are converted into a representation similar to optical flow as motion vector flow($V$) of dimensions $ \in \mathbf{R}^{2\times H \times W}$ , and are resized to match the input image dimensions. Consequently, every pixel within an 8x8 unit shares the identical motion vector value.

\begin{figure}[t]
\begin{center}
  \includegraphics[width=1\linewidth]{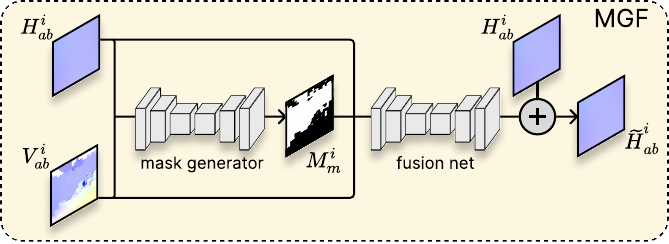}
\end{center}
  \caption{The mask-guided fusion (MGF) module's structure involves utilizing pre-estimated $H_{ab}^i$ and scaled $V_{ab}^i$ to produce a motion rejection mask $M_m^i$. Subsequently, $M^i$, $H^i$ and $V^i$ are input into a fusion network to calculate a residual homography, which is then combined with $H^i_{ab}$ for final fused homography $\widetilde H_{ab}^i$.
  }
  \label{fig_mgfm}
\end{figure}

\subsection{Mask-guided Fusion Module}
\label{MGF}

\begin{figure}[t]
\begin{center}
  \includegraphics[width=1\linewidth]{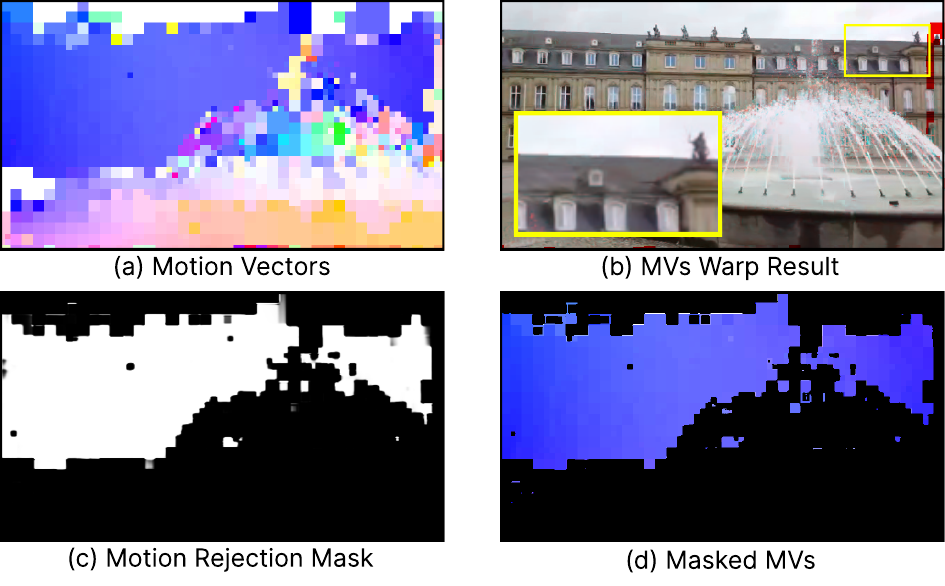}
\end{center}
  \caption{An example of motion vector and motion rejection mask. (a) The visualization flow of MVs. The MVs contains motions in multiple planes. (b) Alignment result of MVs. MVs completed the alignment of the target area with high quality. (c) The motion rejection mask. By applying (c) to (a), motions outside the dominant plane are eliminated as shown in (d). 
  }
  \label{fig_mgf}
\end{figure}

This module is designed to integrate motion vector priors into homography estimation. Homography represents the transformation relative to a dominant plane within the image, while MVs provide a block-wise correspondence that captures varied motions, including dominant plane, foreground, and parallax motions as shown in Fig.~\ref{fig_mgf} (a). Within the regions of the dominant plane, we find that MVs (Fig.~\ref{fig_mgf} (b)) can achieve high quality alignment, but directly using MVs can cause difficulties for network to learn homography. Therefore, we predict a mask to remove useless information from MVs. Remaining useful information is send to the network for fusing.

Inspired by previous work~\cite{pdc}, the MVs-based homography estimation task can be formulated as a probabilistic model:
\begin{equation}
\begin{split}
p(H_{ab} \mid V_{ab}; M_{m}),
\end{split}
\end{equation}
where $p(.)$ is the conditional probability function. Specifically, Motion Rejection Mask $M_{m}$ is designed to evaluate the confidence of similarity at each coordinate in $V_{ab}$ and $H_{ab}$. Matching points with higher confidence are likely to represent dominant plane regions, while those with lower confidence are considered outliers and subsequently filtered out.
To achieve it, $p(.)$ is usually modeled using two conditionally independent Laplace distributions~\cite{pdc} whose density function is given by:
\begin{equation}
    \scriptsize
    \mathcal{L}(H_{ab}|V_{ab}; M_{m})=\prod(\frac{1}{\sqrt{2 \sigma^2}} e^{-\sqrt{\frac{2}{\sigma^2}}\left|u-\mu_u\right|} \cdot \frac{1}{\sqrt{2 \sigma^2}} e^{-\sqrt{\frac{2}{\sigma^2}}\left|v-\mu_v\right|}),
\end{equation}
where the components $(u,v)$ are the offsets in every coordinate of $H_{ab}$ and $(\mu_u, \mu_v)$ are the offsets from $V_{ab}$, and $\sigma^2$ is represented by each coordinate in $M_{m}$. 
An example of $M_{m}$ is as shown in Fig.~\ref{fig_mgf} (c). As Fig.~\ref{fig_mgf} (d) shows, the remaining MVs corresponds to the movement of the dominant plane, which can help homography estimation.

The architecture of our MGF is showcased in Fig.~\ref{fig_mgfm}. For the i-th layer, the homography from the previous layer $H^i_{ab}$ is combined with downscaled MVs $V^i_{ab}$ as the input for the mask generator $\mathcal{G}_m$. This produces a motion rejection mask $M^i_m$ with a single channel, with values ranging from 0 to 1, indicating the similarity between $H^i_{ab}$ and $V^i_{ab}$. Here, values close to 1 represent high similarity. $M^i$ alongside $H^i$ and $V^i$, are then fed into a fusion net $\mathcal{F}_i$ to derive a residual homography that is then added to $H^i_{ab}$. The complete process can be summarized as follows:

\begin{equation}
\label{mas-fusion}
\begin{split}
M^i_m &= \mathcal{G}_m(H_{ab}^i, V^i_{ab}), \\
\widetilde H_{ab}^i &= H_{ab}^i+ \mathcal{F}_i(M^i_m, H_{ab}^i, V^i_{ab}),
\end{split}
\end{equation}
where $\widetilde H_{ab}^i$ is the fused homography flow.

\subsection{Mask-guided Homography Estimation Module}
\label{MGHE}

\begin{figure}[t]
\begin{center}
  \includegraphics[width=1\linewidth]{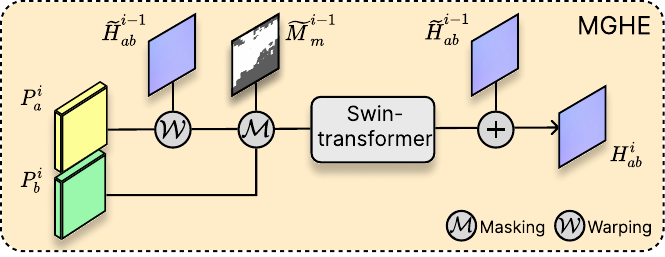}
\end{center}
  \caption{The mask-guided homography estimation (MGHE) module. Adjustment $\widetilde  M^{i-1}_m$ is applied to guide the homography estimation by weighting the $P_b^i$ and warped ${P}_a^{' i}$. 
  }
  \label{fig_mghe}
\end{figure}

To address regions within image features that do not contribute to homography estimation, especially foreground areas, we introduce the novel Mask-Guided Homography Estimation (MGHE) module. In our research, we discovered that the mask $M^{i-1}_m$  applied to filter irrelevant information from the MVs can eliminate interfering factors like foreground motion and extend to removing corresponding interference between features. By processing these refined features, the network can improve its ability to learn homography motion effectively. The architecture of MGHE, as shown in Fig.~\ref{fig_mghe}, employs feature warping, feature filtering, a transformer $\mathcal{T}_i$ from HomoGAN~\cite{homogan} and residual adding.

Initially, mask $M^{i-1}_m$ from the MGF module can be used for its ability to discern between planar and non-planar motion. However, its effectiveness is counting on the similarity between the homography flow $H^{i-1}$ and the motion vector flow $V^{i-1}_{ab}$. In the early layers, $H^{i-1}$ can be noisy, yielding an $M^{i-1}_m$ that may lead to near-zero values, thereby impairing homography estimation. To mitigate this, we adjust $M^{i-1}_m$ using the formula:

\begin{equation}
\label{mask light}
\widetilde{M}^{i-1}_m = M^{i-1}_m \times (1 - \alpha) + \alpha,
\end{equation}
where the adjusted mask, $\widetilde{M}^{i-1}_m$, ranges from $\alpha$ to 1. 

In MGHE module at the i-th layer, the step involves feature sequences $P_a^i$ being warped by $\widetilde H_{ab}^{i-1}$ from the previous layer. Then, both the warped $P_a^i$ and $P_b^i$ are multiplied by $\widetilde{M}^{i-1}_m$ to produce $\hat{P}_a^{'i}$ and $\hat{P}_b^i$, which exclude regions inconsistent with the homography motion. These are subsequently inputted into the transformer. The resulting homography, $H_{ab}^i$, is produced by adding the transformer's output with the current scale factor. The process can be formulated as:

\begin{equation}
\label{mask-esti}
\begin{split}
\hat{P}_a^{' i} &= \mathcal{M}(\mathcal{W}(P_a^i,\widetilde  H_{ab}^{i-1}),\widetilde M^{i-1}_m),\\
\hat{P}_b^i &= \mathcal{M}(P_b^i,\widetilde M^{i-1}_m),\\
H_{ab}^i &= \widetilde H_{ab}^{i-1} + \mathcal{T}_i(\hat{P}_a^{' i}, \hat{P}_b^i),
\end{split}
\end{equation}
where $\mathcal{M}$ is the masking operation.

\subsection{Enhanced Motion Mask}
\label{Mcmc}


As unsupervised learning, the network is trained by minimizing photometric loss in the deep feature space between the warped source and target images. Directly computing the loss for entire image pairs can lead to interference from multiple planes or foreground elements within the image during homography estimation. To enhance the effectiveness of the loss function, we introduce an improvement to the motion rejection mask aimed at better focusing on dominant plane. Our novel concept involves the creation of an enhanced motion mask, denoted as $M_{e}$, designed specifically for loss computation. Illustrated in Fig.~\ref{fig_pip}, our approach integrates two key components: the coplanarity-aware mask and the motion-rejection mask.

The coplanarity-aware mask $M_c$ leverages foundational principles from HomoGAN~\cite{homogan}, identifying predominant planar regions. Differently, $M_c$ is uniquely derived from warped feature $F_{a}^{'}$ (warping $F_{a}$ by $H_{ab}$) and $F_b$, offering a distinctive approach to plane delineation. Additionally, the motion-rejection mask $M_m$ is computed utilizing the homography, $H_{ab}$, obtained through a coars-to-fine process which represents the finest result.

By focusing on different aspects of the image, $M_c$ on feature disparities and $M_m$ on motion discrepancies, we achieve a comprehensive outlier filtration approach. The combination of these masks through element-wise multiplication further highlight the focus areas. Specifically, regions identified by both masks as belonging to the dominant plane are emphasized. This process is encapsulated by the following formula:

\begin{equation}
\label{mask}
M_{e} = \mathcal{G}_c(F_{a}^{'}, F_b) \cdot \mathcal{G}_m(H_{ab}, V_{ab}),
\end{equation}
where $\cdot$ is element-wise product, $\mathcal{G}_c$ is the mask generator for the coplanarity-aware mask and $\mathcal{G}_m$ is the mask generator for the motion rejection mask.

\subsection{Loss Function}

In the unsupervised learning process, several loss functions are minimized for the training of the homography estimation network. For simplicity, we will only discuss the forward loss in this section. Firstly, we compute an alignment loss  $\ell_{align}$ by comparing feature maps before and after warping by the predicted homographies. The distance $\ell_{tri}$ is obtained by the pixel-wise triplet loss~\cite{cahomo} as:
\begin{equation}
\label{Gap}
\ell_{tri} = {\rm max}(||F_{a}^{'} - F_b||_1-||F_{a} - F_b||_1 + 1,0).
\end{equation}
We further apply the predicted masks to $\ell_{tri}$, emphasizing the dominant plane to compute $\ell_{align}$ as follows:
\begin{equation}
\label{L align}
\ell_{align} = \frac{\sum_i M_{e}\ell_{tri}}{\sum_i M_{e}},
\end{equation}
where i is the pixel index.

The second term is the feature identity loss  $\ell_{FIL}$ that ensures
the feature projector $\mathcal{P}$ to be warp-invariant [35], which is given by:
\begin{equation}
\label{L fil}
\begin{split}
    \ell_{FIL} =&||\mathcal{W}(H_{ab}, \mathcal{P}(I_{a})) - \mathcal{P}(\mathcal{W}(H_{ab}, (I_{a}))||_1.
\end{split}
\end{equation}
This loss encourages $\mathcal{P}$ to be consistent through geometric transformations while filtering out luminance variations.

For the plane estimation, $\ell_{plane}$ comprises two components, we firstly apply the negative log likelihood loss:

\begin{equation}
    \ell_{nll} = -\log p(H_{ab} \mid V_{ab}; M_{e}).
\end{equation}

Additionally, to avoid an all-zero mask, we employ a binary cross-entropy loss between the masks and an all-ones mask:
\begin{equation}
\label{L bce}
    \ell_{bce} = {\rm BCE} (M_{e}, 1).
\end{equation}

The overall loss for plane estimation is:
\begin{equation}
\label{L plane}
    \ell_{plane} = \ell_{nll} + 0.05 \ell_{bce}.
\end{equation}

The total loss function is therefore:
\begin{equation}
\label{L total}
    \ell_{total} = \ell_{align} + \ell_{FIL} + \ell_{plane}.
\end{equation}

\section{Experiment}

\begin{figure*}[!h]
\begin{center}
  \includegraphics[width=1\linewidth]{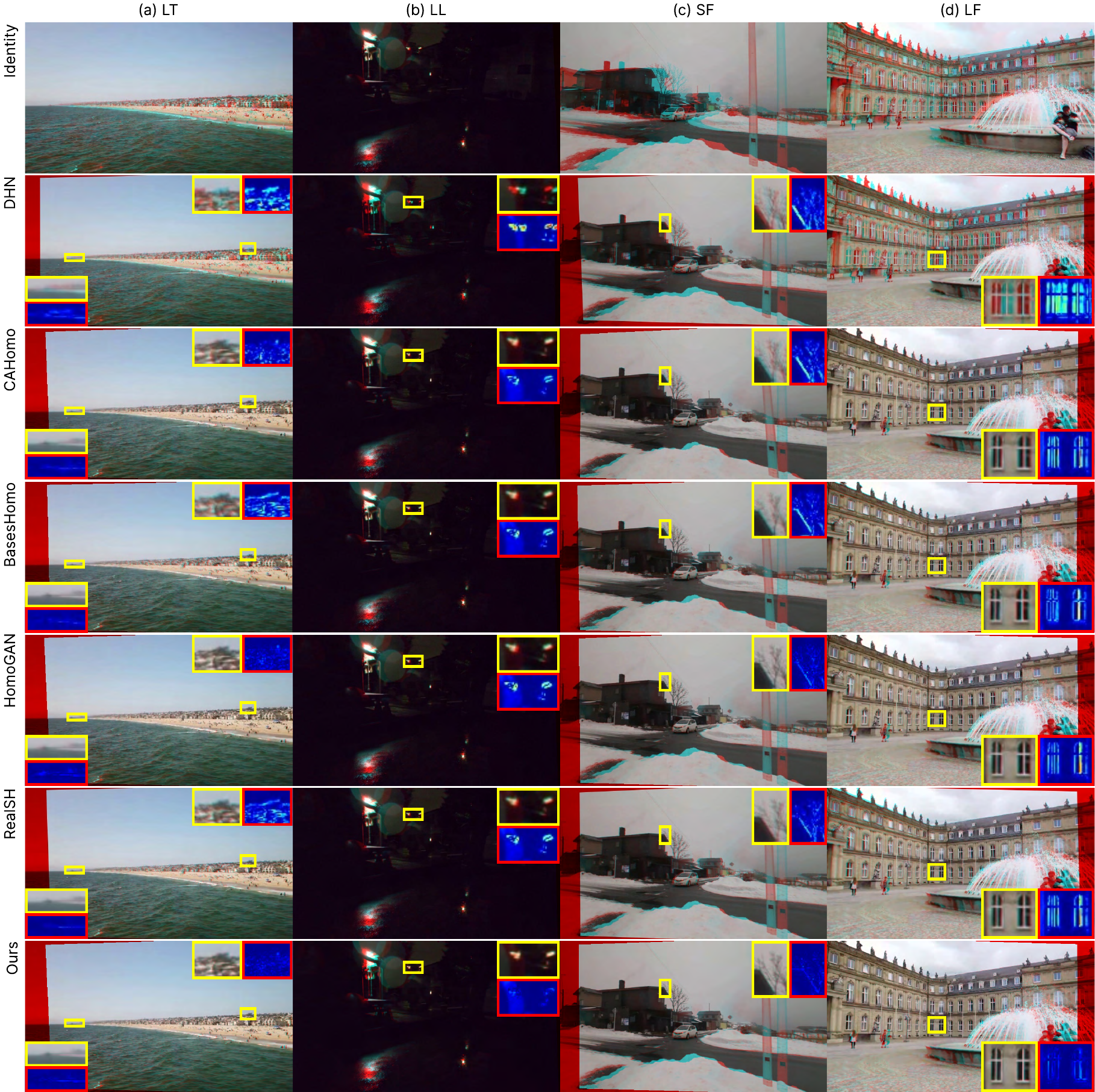}
\end{center}
  \caption{Qualitative results of our method and other existing learning-based methods on the CA-unsup\cite{cahomo} test set. The images are generated by superimposing the warped source images on the target image. Error-prone regions are highlighted with yellow boxes, and the red boxes show the content difference between the two images in the error-prone regions. Best viewed by zooming in.}
  \label{fig_CA_learning}
\end{figure*}

\begin{figure*}[!h]
\begin{center}
  \includegraphics[width=1\linewidth]{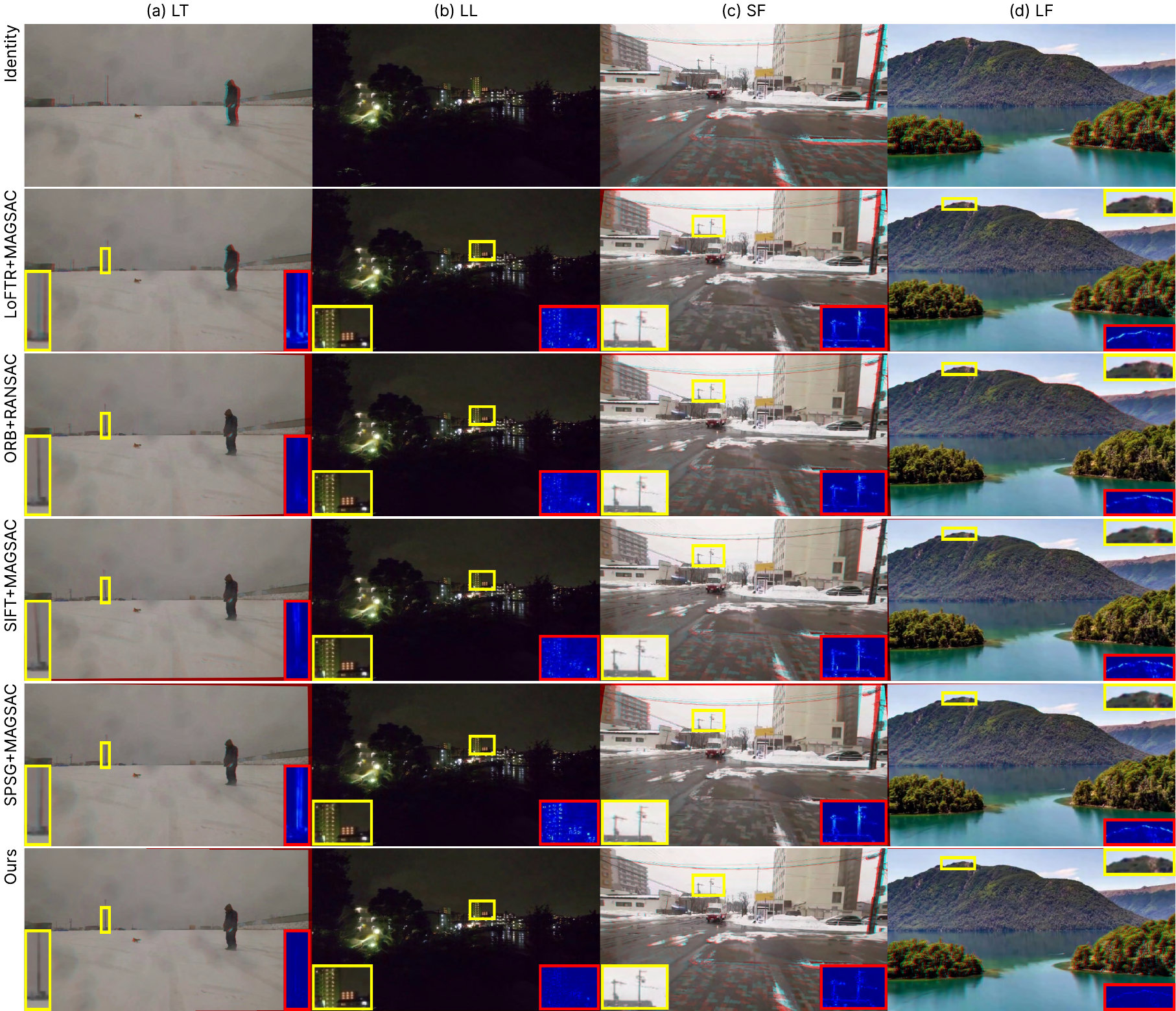}
\end{center}
  \caption{Qualitative results of feature-based methods and our method. For each feature-based method, we show its results with the best performed outlier rejection algorithm. Best viewed by zooming in.}
  \label{fig_CA_fea}
\end{figure*}

\begin{figure*}[!h]
\begin{center}
  \includegraphics[width=1\linewidth]{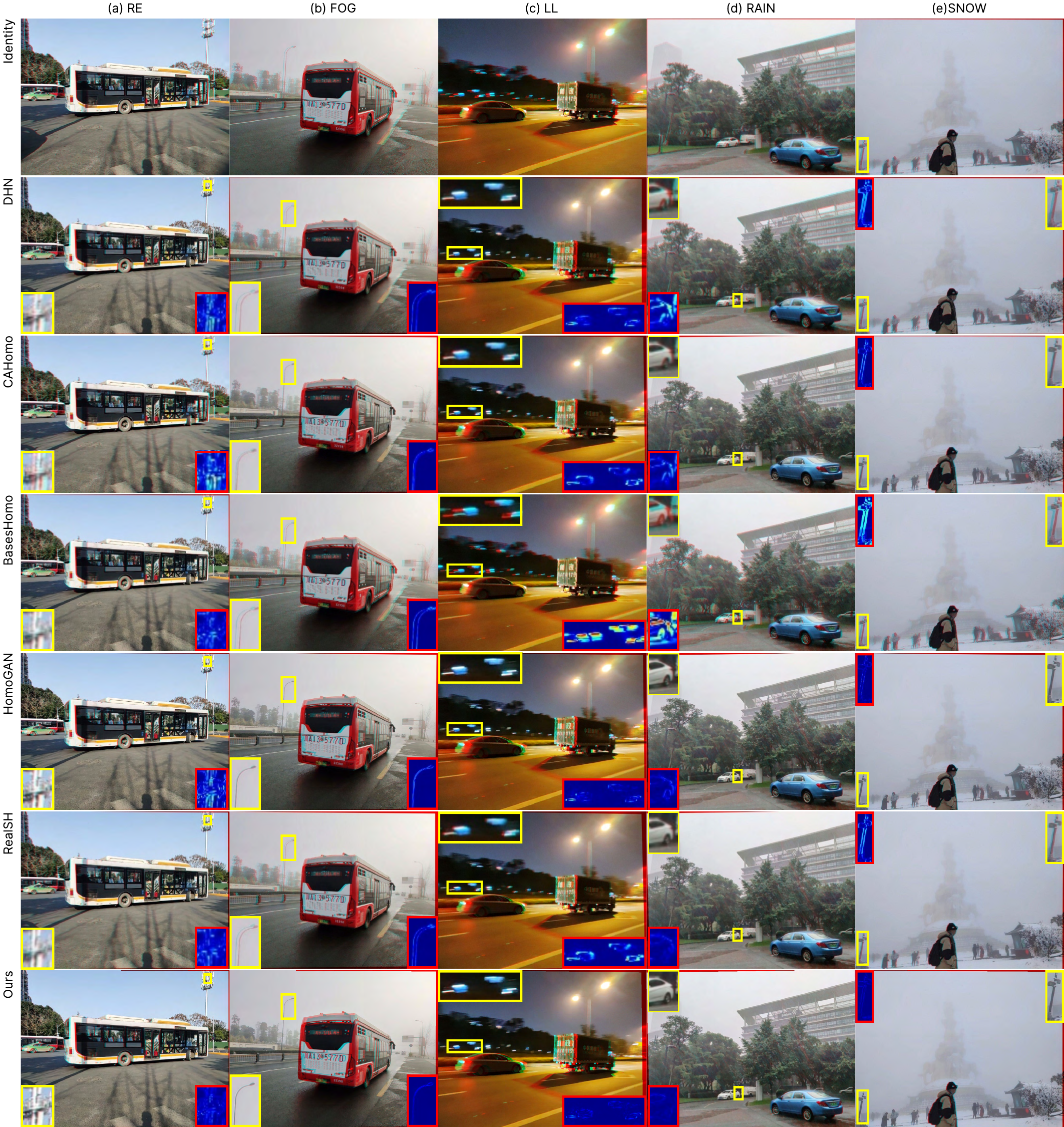}
\end{center}
  \caption{Qualitative results of our method and other competitive methods on the GHOF~\cite{gyroflow+} test set. Best viewed by zooming in.}
  \label{fig_GHOF}
\end{figure*}

\begin{table*}
    \begin{center}
    \resizebox{0.98\linewidth}{!}{
    \begin{tabular}{rl|lllllll}
    \toprule
    1) & Methods & \multicolumn{1}{c}{AVG} & \multicolumn{1}{c}{RE} & \multicolumn{1}{c}{LT} & \multicolumn{1}{c}{LL} & \multicolumn{1}{c}{SF} & \multicolumn{1}{c}{LF} \\
    \midrule
    2) & $\mathcal{I}_{3\times3}$ & 6.70 ($\pm75.23$) & 7.75 ($\pm39.63$) & 7.65 ($\pm91.51$) & 7.21 ($\pm76.71$) & 7.53 ($\pm136.49$) & 3.39 ($\pm43.84$) \\
    \midrule
    3) & SIFT~\cite{sift} + RANSAC~\cite{ransac} & 1.41 ($\pm2.79$) & 0.30 ($\pm1.22$) & 1.34 ($\pm4.37$) & 4.03 ($\pm3.99$) & 0.81 ($\pm1.14$) & 0.57 ($\pm0.97$) \\
    4) & SIFT~\cite{sift} + MAGSAC~\cite{magsac} & 1.34 ($\pm2.70$) & 0.31 ($\pm1.20$) & 1.72 ($\pm4.47$) & 3.39 ($\pm3.61$) & 0.80 ($\pm1.08$) & 0.47 ($\pm0.97$) \\
    5) & ORB~\cite{orb} + RANSAC~\cite{ransac} & 1.48 ($\pm3.57$) & 0.85 ($\pm2.57$) & 2.59 ($\pm6.23$) & 1.67 ($\pm3.77$) & 1.10 ($\pm1.75$) & 1.24 ($\pm1.06$) \\
    6) & ORB~\cite{orb} + MAGSAC~\cite{magsac} & 1.69 ($\pm3.47$) & 0.97 ($\pm2.01$) & 3.34 ($\pm6.03$) & 1.58 ($\pm3.90$) & 1.15 ($\pm1.77$) & 1.40 ($\pm1.25$) \\
    7) & SPSG~\cite{superpoint, superglue} + RANSAC~\cite{ransac} & 0.71 ($\pm1.00$) & 0.41 ($\pm0.42$) & 0.87 ($\pm1.76$) & 0.72 ($\pm0.65$) & 0.80 ($\pm0.69$) & 0.75 ($\pm0.94$) \\
    8) & SPSG~\cite{superpoint, superglue} + MAGSAC~\cite{magsac} & 0.63 ($\pm1.03$) & 0.36 ($\pm0.33$) & 0.79 ($\pm1.75$) & 0.70 ($\pm0.62$) & 0.71 ($\pm0.70$) & 0.70 ($\pm1.13$) \\
    9) & LoFTR~\cite{loftr} + RANSAC~\cite{ransac} & 1.44 ($\pm2.53$) & 0.56 ($\pm0.73$) & 2.70 ($\pm5.20$) & 1.36 ($\pm1.20$) & 1.05 ($\pm1.28$) & 1.52 ($\pm1.12$) \\
    10) & LoFTR~\cite{loftr} + MAGSAC~\cite{magsac} & 1.39 ($\pm2.45$) & 0.55 ($\pm0.69$) & 2.57 ($\pm5.09$) & 1.33 ($\pm1.04$) & 1.05 ($\pm1.17$) & 1.41 ($\pm1.10$) \\
    \midrule
    11) & DHN~\cite{dhn} & 2.87 ($\pm4.62$) & 1.51 ($\pm3.16$) & 4.48 ($\pm6.12$) & 2.76 ($\pm5.42$) & 2.62 ($\pm3.44$) & 3.00 ($\pm4.23$) \\
    12) & LocalTrans~\cite{localtrans} & 4.21 ($\pm5.94$) & 4.09 ($\pm3.00$) & 4.84 ($\pm6.99$) & 4.55 ($\pm6.20$) & 5.30 ($\pm8.27$) & 2.25 ($\pm3.43$) \\
    13) & IHN~\cite{ihn} & 4.67 ($\pm6.04$) & 4.85 ($\pm4.44$) & 5.54 ($\pm7.36$) & 5.10 ($\pm5.99$) & 5.04 ($\pm7.25$) & 2.84 ($\pm4.50$)\\
    14) & RealSH~\cite{realsh} & \underline{0.34 ($\pm0.39$)} & \underline{0.22 ($\pm0.20$)} & \underline{0.35 ($\pm0.38$)} & \underline{0.44 ($\pm0.49$)} & \underline{0.42 ($\pm0.39$)} & \underline{0.29 ($\pm0.37$)} \\
    \midrule
    15) & CAHomo~\cite{cahomo} & 0.88 ($\pm0.54$) & 0.73 ($\pm0.22$) & 1.01 ($\pm0.58$) & 1.03 ($\pm0.69$) & 0.92 ($\pm0.46$) & 0.70 ($\pm0.60$) \\
    16) & BasesHomo~\cite{baseshomo} & 0.50 ($\pm0.45$) & 0.29 ($\pm0.24$) & 0.54 ($\pm0.41$) & 0.65 ($\pm0.53$) & 0.61 ($\pm0.55$) & 0.41 ($\pm0.72$) \\
    17) & HomoGAN~\cite{homogan} & 0.39 ($\pm0.42$) & 0.22 ($\pm0.17$) & 0.41 ($\pm0.32$) & 0.57 ($\pm0.62$) & 0.44 ($\pm0.47$) & 0.31 ($\pm0.37$)\\
    18) & Ours & \textbf{0.31 ($\pm0.35$)} & \textbf{0.18 ($\pm0.14$)}& \textbf{0.29 ($\pm0.25$)} & \textbf{0.42 ($\pm0.52$)} & \textbf{0.37 ($\pm0.33$)} & \textbf{0.26 ($\pm0.37$)}\\
    
   \bottomrule
   \end{tabular}}
\end{center}

\caption{The point matching errors (PME) of our method and all comparison methods on the CA-unsup~\cite{cahomo} test set. The best and second-best results are highlighted in \textbf{bold} and \underline{underlined}. The values in parentheses represent the standard deviation of a specific category. SPSG indicates SuperPoint with SuperGlue.}
\label{tab:CAunsup}
\end{table*}

\begin{table}
    \begin{center}
    \resizebox{0.98\linewidth}{!}{
    \begin{tabular}{rl|lllllll}
    \toprule
    1) & Methods & \multicolumn{1}{c}{AVG} & \multicolumn{1}{c}{RE} & \multicolumn{1}{c}{FOG} & \multicolumn{1}{c}{LL} & \multicolumn{1}{c}{RAIN} & \multicolumn{1}{c}{SNOW} \\
    \midrule
    2) & $\mathcal{I}_{3\times3}$ & 6.33 & 4.94  & 7.24 & 8.09 & 5.48 & 5.89 \\
    \midrule
    3) & SIFT~\cite{sift} + MAGSAC~\cite{magsac} & 3.90 & \textbf{0.64} & 4.01 & 9.77 & 0.70 & 4.40 \\
    4) & ORB~\cite{orb} + RANSAC~\cite{ransac} & 15.14 & 6.92 & 31.27 & 27.80 & 1.82 & 7.90 \\
    5) & SPSG~\cite{superpoint,superglue} + MAGSAC~\cite{magsac} & 3.28 & 3.41 & 1.46 & 7.61 & 0.75 & 3.91 \\
    6) & LoFTR~\cite{loftr} + MAGSAC~\cite{magsac} & 2.55 & 2.76 & 1.16 & 5.34 & 0.52 & 2.98 \\
    \midrule
    7) & DHN~\cite{dhn} & 6.61 & 6.04 & 6.02 & 7.68 & 6.99 & 6.32 \\
    8) & LocalTrans~\cite{localtrans} & 5.72 & 4.06 & 6.49 & 5.95 & 5.78 & 6.34 \\
    9) & IHN~\cite{ihn} & 8.17 & 7.10 & 8.71 & 9.34 & 6.57 & 9.13 \\
    10) & RealSH~\cite{realsh} & \underline{1.72} & 1.60  & 0.88 & 4.42 & \underline{0.43} & \textbf{1.28} \\
    \midrule
    11) & CAHomo~\cite{cahomo} & 3.87 & 4.10 & 3.84 & 6.99 & 1.27 & 3.17\\
    12) & BasesHomo~\cite{baseshomo} & 2.28 & 2.02 & 1.43 & 4.90 & 0.78 & 2.29 \\
    13) & HomoGAN~\cite{homogan} & 1.95 & 1.73 & \underline{0.60} & \underline{3.95} & 0.47 & 3.02 \\
    14) & Ours & \textbf{1.21} & \underline{0.85}& \textbf{0.31} & \textbf{2.91} & \textbf{0.38 } & \underline{1.58}\\
    
   \bottomrule
   \end{tabular}}
\end{center}
\caption{The point matching errors (PME) of our method and all comparison methods on the GHOF~\cite{gyroflow+} test set. The best and second-best results are highlighted in \textbf{bold} and \underline{underlined}.}
\label{tab:Compare_GHOF}
\end{table}

\subsection{Dataset}

\textbf{CA-unsup~\cite{cahomo}}: The CA-unsup dataset is dedicating for deep homography task, which comprises 442,000 training pairs and 4,200 testing pairs with an image resolution of 320 × 640 pixels. We augmented both the training and testing sets with the corresponding motion vectors (MVs). The images in each subset are distributed among five scene categories: regular (RE), low texture (LT), low light (LL), small foreground (SF), and large foreground (LF). Notably, the latter four categories are considered challenging environments for accurate homography estimation. For evaluation purposes, each testing image is equipped with six 6 pairs of ground-truth matching points. The primary evaluation metric employed is the average L2 distance between the model's predicted points and these ground-truth reference points on the target images.

\textbf{GHOF~\cite{gyroflow+}}:The GHOF dataset is specifically designed to facilitate the training and evaluation of gyroscope-based optical flow and homography estimation algorithms. It includes a training set of approximately 10,000 video frames taken with a non-OIS (non-Optical Image Stabilizer) camera phone across diverse environmental conditions and seasons such as standard lighting, low light, fog, rain, and snow. These particular conditions were selected for their complexity and the presence of veiling and streak effects, which pose significant capture challenges. Moreover, the dataset features scenarios that incorporate parallax variations and dynamic human movement to address challenges associated with rotational camera motion and non-rigid motion within the scene. The evaluation set, on the other hand, consists of 246 image pairs annotated with sparse correspondences for a robust assessment of performance. The key metric for evaluating models on this dataset is the average L2 distance between the predicted points by the model and the annotated ground-truth points on the target images. 

\subsection{Implementation Details}

During the training phase, we augment the dataset by randomly cropping patches of size 384×512 from near the center of the original images. This ensures that the coordinates remain within bounds after warping. We have set the number of scale levels to three. Our framework consists of 2.07 million parameters, with the network requiring a total of 46.17 GFLOPs, of which 5.39 GFLOPs are allocated to the fusion part. Our network is implemented using PyTorch and trained on a cluster of four NVIDIA RTX 3090 GPUs. We utilize the Adam optimizer~\cite{adam} with a starting learning rate of $1 \times 10^{-4}$, which decays by a factor of 0.8 after each epoch. The training process is divided into two stages: the first stage consists of 10 epochs with a batch size of 24, followed by a second stage of 2 epochs during which the learning rate is reduced to $4 \times 10^{-6}$. This reinitialization helps facilitate finer adjustments during the latter part of the training process.

\subsection{Comparison with Existing Method}

\subsubsection{Comparison Methods.}

We compare our method with three categories of existing homography estimation methods: 1) feature-based methods including SIFT~\cite{sift}, ORB~\cite{orb}, SuperPoint~\cite{superpoint} with SuperGlue (SPSG)~\cite{superglue}, and LoFTR~\cite{loftr}, 2) supervised methods including DHN~\cite{dhn}, LocalTrans~\cite{localtrans}, IHN~\cite{ihn} and RealSH~\cite{realsh}, 3) unsupervised methods including CAHomo~\cite{cahomo}, BasesHomo~\cite{baseshomo} and HomoGAN~\cite{homogan}. For feature-based methods, two different outlier rejection algorithms are applied as RANSAC~\cite{ransac} and MAGSAC~\cite{magsac} respectively. All learning-based methods are trained on the CA-sup dateset.

\subsubsection{Quantitative Comparison}

In Table~\ref{tab:CAunsup}, we present the quantitative results for all methods under comparison. We employ the points matching error (PME) as the metric to assess the average geometric distance between homography-warped source points and ground-truth target points within dominant plane regions for each pair of test images. Besides computing the mean PME, standard deviation serves as another commonly used second-order statistic to assess robustness ensures that comparisons are statistically significant. Standard deviation indicates how spread out method values are from their mean, offering insights into PME dispersion and method stability.

The results from Table~\ref{tab:CAunsup} illustrate that our method surpasses all others on the dataset, achieving state-of-the-art performance. Notably, it demonstrates an 8.82\% improvement over the previously best method, RealSH, reducing the matching error from 0.34 to 0.31. In regular (RE) scenes, both learning-based methods and feature-based methods typically excel, owing to the images' abundant features. With the inclusion of prior MVs, our model achieves an 18.18\% lower error in this category compared to RealSH and HomoGAN. The small foreground (SF) and large foreground (LF) scene categories often present challenges due to dynamic contents and the presence of multiple planes. Our method outshines other learning-based methods—including those armed with outlier rejection mechanisms such as CA-Unsupervised, BasesHomo, and HomoGAN-and exhibits superior performance in managing these outliers. Moreover, in scenes with low light (LL) and low texture (LT), our outlier rejection mechanism remains effective by generating more precise masks, thereby maintaining focus on the substantially smaller domain plane. In these challenging conditions, our method reduces errors by at least 17.14\% for LT and 4.55\% for LL, outperforming both supervised and unsupervised methods.

Our method has the lowest standard deviation in AVG, RE, LT, SF, and LF, and the second lowest in LL, demonstrating excellent stability with the utilization of prior MVs. Additionally, we conducted a t-test comparing our method to the best existing method, realSH~\cite{realsh}. The resulting t-value of 3.7102, which exceeds the common threshold of 2, indicates that the difference between our results and the previous sota method is statistically significant.

\textbf{Generalization experiment}:
We conducted further evaluations on feature-based methods that performed well on the CA-unsup dataset, learning-based methods, and our own method using the GHOF test set. To validate our method, we added the corresponding MVs to the test set (GHOF). As shown in Table~\ref{tab:Compare_GHOF}, our method achieves the best results in FOG, LL, and RAIN scenes and the second-best results in RE and SNOW scenes, resulting in a 29.65\% improvement in average point matching errors from 1.72 to 1.21. Notably, our method excelled in FOG and RAIN scenes, which were not included in the training dataset (CA-unsup), thus demonstrating the effectiveness of our MVs prior and MGF module.

\subsubsection{Qualitative Comparison}

We provide a qualitative comparison between our method, CodingHomo, and six deep learning-based methods (two supervised and four unsupervised) in four challenging scenes of the CA-unsup test set. In Fig.~\ref{fig_CA_learning} (a), the scene presents challenges such as low texture and distant buildings. Fig.~\ref{fig_CA_learning} (b) is a low light scene with a dynamic foreground building. Fig.~\ref{fig_CA_learning} (c) is challenging due to the relatively small portion of the image occupied by the plane of interest and the presence of a telephone pole as an interrupting foreground element. Additionally, the large fountain in Fig.~\ref{fig_CA_learning} (d) results in significant depth disparity from the foreground to the background. As highlighted in the red and yellow boxes, existing methods cannot align these images as well as ours. 

\begin{figure*}[ht]
\begin{center}
  \includegraphics[width=1\linewidth]{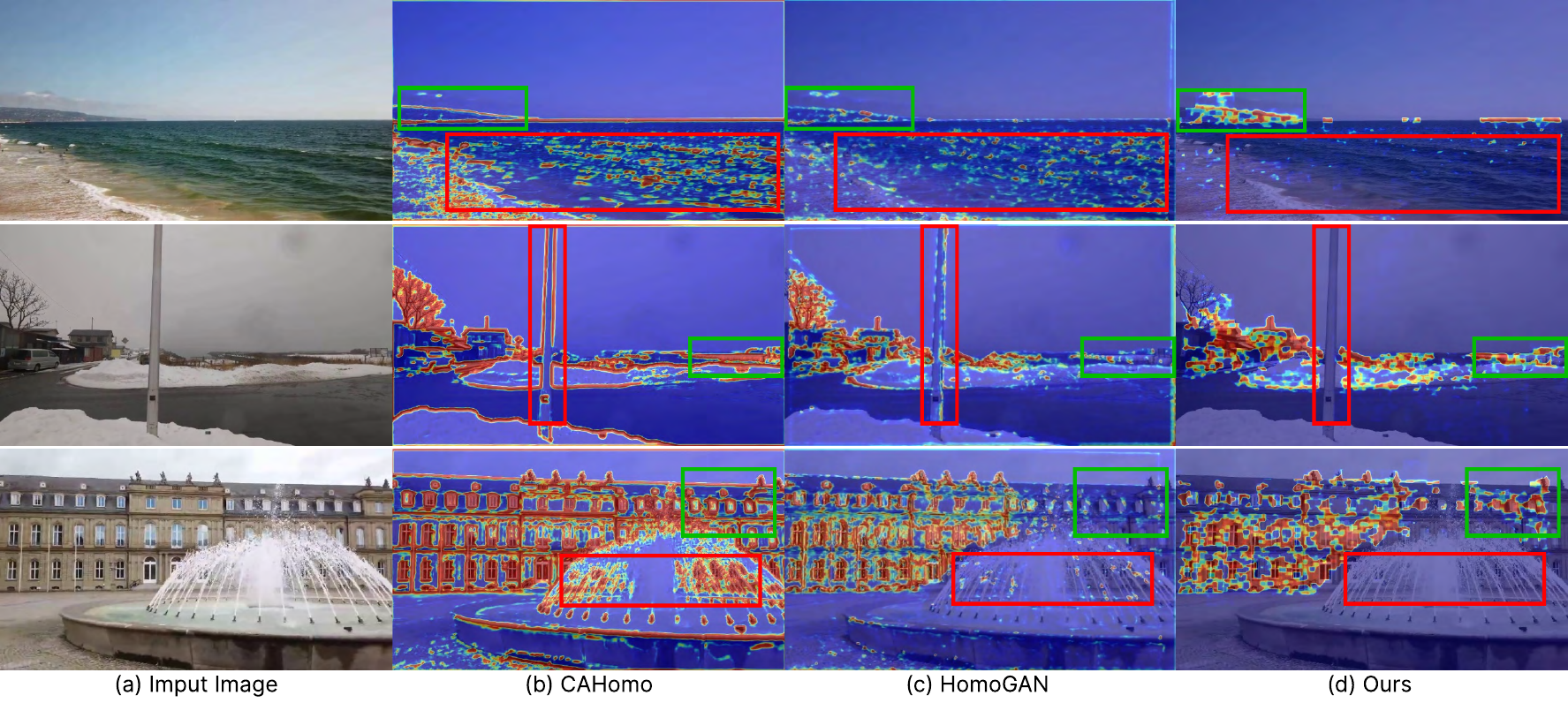}
\end{center}
  \caption{Visualization results of outlier rejection masks of our method and other competitive methods on the CA-unsup~\cite{cahomo} test set. Red Blocks indicate outlier region. Green Blocks donates inlier region. }
  \label{fig_vis_mask}
\end{figure*}

In Fig.~\ref{fig_CA_fea}, we compared our methods with feature-based approaches on the CA-unsup test set. In Fig.~\ref{fig_CA_fea} (a) and Fig.~\ref{fig_CA_fea} (b), feature-based methods failed due to insufficient correspondence in low-texture and low-light scenes. Despite applying outlier rejection methods like RANSAC~\cite{ransac} and MAGSAC~\cite{magsac}, these feature-based methods still struggled in scenes with moving objects, as illustrated in Fig.~\ref{fig_CA_fea} (c) and Fig.~\ref{fig_CA_fea} (d), resulting in blurry boundaries. In contrast, our method remains robust in these scenarios without relying on keypoints.

\textbf{Generalization experiment}: 
In Fig.~\ref{fig_GHOF}, we present the comparison results of our method with deep learning-based approaches on the GHOF test set, including scenes with FOG, RAIN, and SNOW depicted in Fig.~\ref{fig_GHOF} (b), Fig.~\ref{fig_GHOF} (d), and Fig.~\ref{fig_GHOF} (e), respectively. These scenes were not included in our training set. Fig.~\ref{fig_GHOF} (a), Fig.~\ref{fig_GHOF} (b), and Fig.~\ref{fig_GHOF} (c) is challenges due to dynamic vehicles. As the Our method outperformed others by leveraging effective prior information from MVs.

\subsection{Robustness evaluation}

\begin{figure*}[t]
\begin{center}
  \includegraphics[width=1\linewidth]{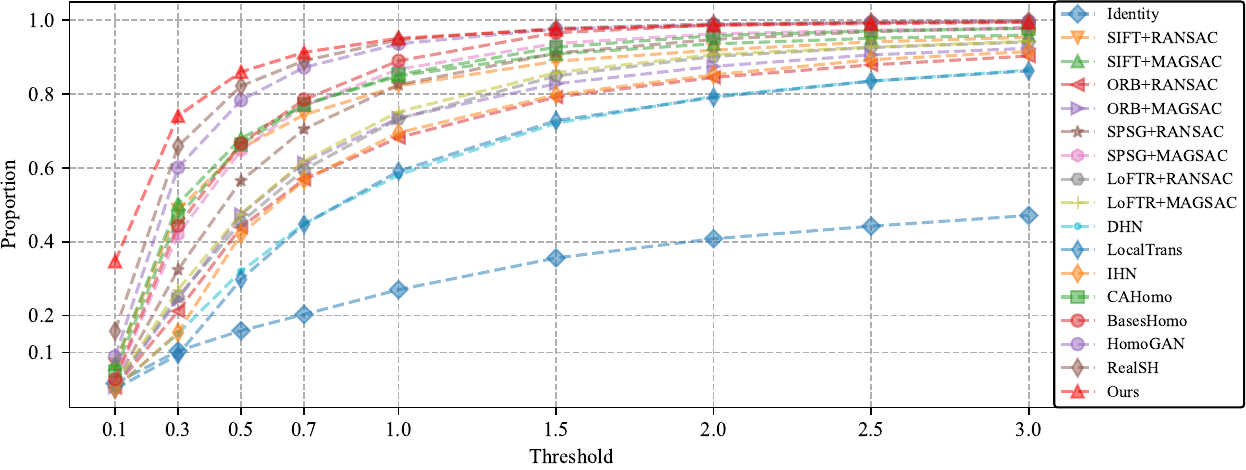}
\end{center}
  \caption{The proportion of inliers of our method and all comparison methods under various thresholds. Inliers indicate points with errors under the threshold. The higher position of curves represents better robustness.}
  \label{fig_rub}
\end{figure*}

Furthermore, we assessed the robustness of all comparison methods on the CA-unsup test set by setting thresholds to calculate the proportion of inlier predictions. Fig.~\ref{fig_rub} illustrates a series of curves where the X-axis ranges from 0.1 to 3.0, representing the threshold values, and the Y-axis ranges from 0.0 to 1.0, representing the percentage of inliers. A higher position of the curves indicates better robustness. Our method achieved an inlier percentage that is 18.9\% higher than the second-best method (34.7\% vs. 15.8\%) with a threshold of 0.1. With a threshold of 0.3, our inlier percentage 8.2\% higher than the second-best method (74.1\% vs. 65.9\%). The significant increase demonstrated the robustness of our method.

\subsection{Discussion}
Based on previous experiment result, a comparison with previous sota methods are as follows: 

Supervised methods DHN~\cite{dhn} and RealSH~\cite{realsh} fails as lack of direct guidance to the dominant plane and unreal artifacts. Our method outperforms unsupervised methods like CAHomo~\cite{cahomo}, BasesHomo~\cite{baseshomo}, and HomoGAN~\cite{homogan}, which have outlier rejection mechanisms, in accurately focusing on the dominant plane.  BasesHomo~\cite{baseshomo} relies on implicit constraints that lack an intuitive impact on the loss function, thereby limiting its performance. On the other hand, CAHomo~\cite{cahomo} suffers from a lack of direct constraints in masks, leading to an excessive focus on object edges during mask learning, especially challenging when dealing with multiple planes. Additionally, HomoGAN~\cite{homogan} faces training difficulty and mode collapse due to its use of GAN loss and tends to rely heavily on image features, resulting in erroneous rejections, particularly concerning larger planes. Our approach effectively generate the mask by modeling the confidence of similarity between the prior MVs and estimated homography though nll loss. The visualization comparison presented in Fig.~\ref{fig_vis_mask} showcases the differences in masks generated by CAHomo~\cite{cahomo}, HomoGAN~\cite{homogan}, and our method. Our masks can eliminate outlier effectively while less wrong rejection.

Besides, pure learning-based methods exhibiting notable performance degradation in scenarios beyond those covered in the dataset, especially supervised methods including DHN~\cite{dhn},  IHN~\cite{ihn} and LocalTrans~\cite{localtrans}. MVs offer a unique matching relationship between images, distinct from homography, serving as a valuable reference for the network. As a widely utilized traditional algorithm, MVs exhibit stability and strong generalization properties, thereby augmenting network performance in robustness and generalization effectively.

\begin{figure*}[!ht]
\begin{center}
  \includegraphics[width=1\linewidth]{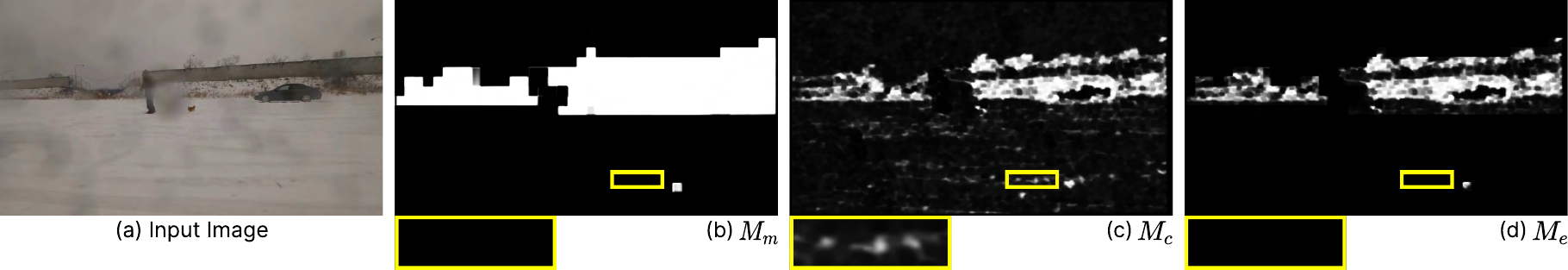}
\end{center}
  \caption{ Illustration of masks. The yellow box indicates that an interrupted area is finally eliminated in the fused masks.}
  \label{fig_masks}
\end{figure*}

\subsection{Ablation Studies}

\begin{figure}[t]
\begin{center}
  \includegraphics[width=1\linewidth]{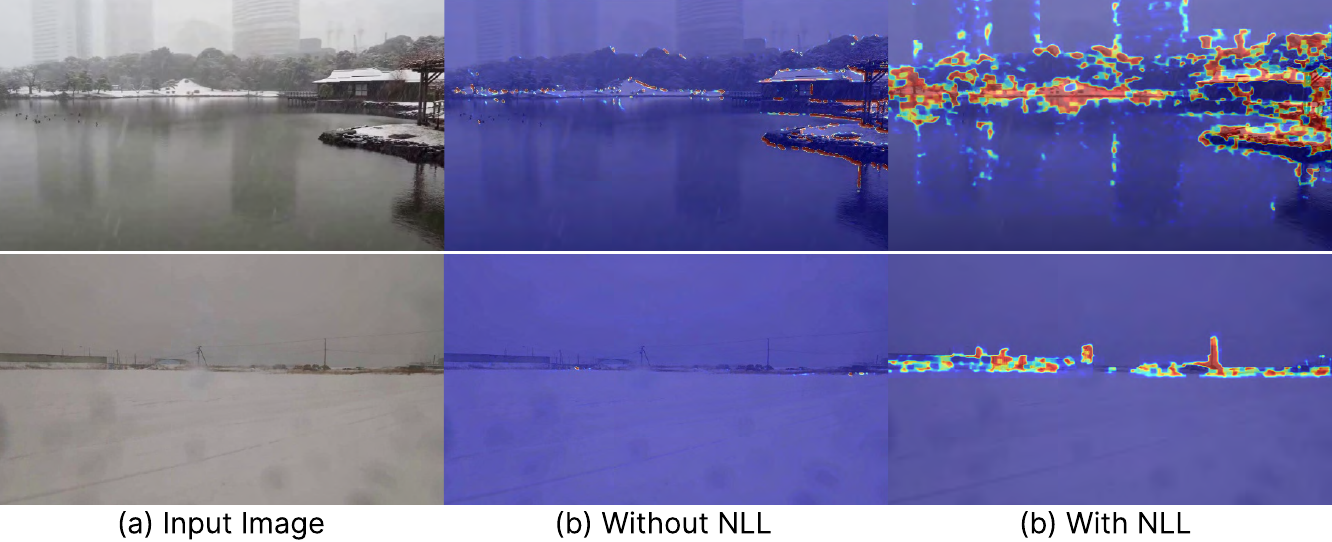}
\end{center}
  \caption{Illustration of the masks generated without/with NLL constraint. Masks without NLL constraint tend to have excessively small attention areas, impacting correct alignment, whereas masks with NLL constraint can effectively focus on the dominant plane.}
  \label{fig_nll}
\end{figure}

We perform several ablation studies on different components to gain insight into the functionality of these main modules. Fig.~\ref{fig_masks}, Fig.~\ref{fig_nll}, Table~\ref{tab_alpha}, and Table~\ref{tab:Ablation Studies} display the ablation results.

\begin{table}[t]
    \begin{center}
    \resizebox{0.98\linewidth}{!}{
    \begin{tabular}{rl|lllllll}
    \toprule
    1) & modification & \multicolumn{1}{c}{AVG} & \multicolumn{1}{c}{RE} & \multicolumn{1}{c}{LT} & \multicolumn{1}{c}{LL} & \multicolumn{1}{c}{SF} & \multicolumn{1}{c}{LF} \\
    \midrule
    2) & w/o guide mask  & 0.36 & 0.20 & 0.35 & 0.50 & 0.42 & 0.31 \\
    3) & w/o mask in loss & 0.46 & 0.23 & 0.47 & 0.70 & 0.54 & 0.35 \\
    4) & coplanarity-aware mask & 0.33 &0.19 &0.31 & 0.45 & 0.41 & 0.30 \\
    5) & motion-rejection mask & 0.33 & 0.19 & 0.30 & 0.45 & 0.41 & 0.29 \\
    6) & w/o NLL loss: & 0.39 & 0.19 & 0.46 & 0.54 & 0.41 & 0.29\\  
    \midrule
    7) & Ours & 0.31 & 0.18 & 0.29 & 0.42 & 0.37 & 0.26 \\
    \bottomrule
    \end{tabular}}
    \end{center}
    \caption{ Results of ablation studies. Please refer to the text for details. w/o donates without. }
    \label{tab:Ablation Studies}
\end{table}

\noindent \textbf{Mask Guided Fusion}: 
    As described in Sec.~\ref{MGF}, we propose a motion reject mask to eliminate the interrupt motion within the $V_{ab}$ before fusing $H_{ab}$ and $V_{ab}$. In this experiment, we excluded the mask from the MGF module and directly fused $H_{ab}$ with $V_{ab}$. The results are presented in Table~\ref{tab:Ablation Studies} . A comparison between these results and our final approach reveals an degrade in error from 0.31 to 0.36. This shift underscores the significance of employing a motion reject mask $M_{m}$, showcasing its role in effectively suppressing interference and extracting relevant segments from $V_{ab}$ as correctly constructing the probabilistic model to distinguish between inliers and outliers.

\begin{table}[t]
    \begin{center}
    \resizebox{0.98\linewidth}{!}{
    \begin{tabular}{rl|lllllll}
    \toprule
    1) & modification & \multicolumn{1}{c}{AVG} & \multicolumn{1}{c}{RE} & \multicolumn{1}{c}{LT} & \multicolumn{1}{c}{LL} & \multicolumn{1}{c}{SF} & \multicolumn{1}{c}{LF} \\
    \midrule
    2) & 1.0 & 0.34 & 0.19 & 0.31 & 0.45 & 0.44 & 0.31 \\
    3) & 0.4 & 0.32 & 0.19 & 0.29 & 0.42 & 0.40 & 0.29 \\ 
    4) & 0.2 & 0.31 & 0.18 & 0.29 & 0.42 & 0.37 & 0.26 \\
    5) & 0.0 & 0.38 & 0.19 & 0.46 & 0.54 & 0.41 & 0.29 \\
    
    \bottomrule
    \end{tabular}}
    \end{center}
    \caption{Result of ablation studies with different $\alpha$.}
    \label{tab_alpha}
\end{table}

\noindent \textbf{MGHE:} 
    In Sec.~\ref{MGHE}, we extending the mask used in the fusion module to the estimation module. In this experiment, we tested different thresholds of $\alpha$ from 0, fully using, to 1, not using. The results are presented in Table~\ref{tab_alpha}, by comparing row 2 and row 4, we observed that appropriately using masks for guiding homography estimation improving the performance from 0.34 to 0.31. However, fully using masks in row 5 leads to undesired increases in errors especially in LT and LL scenes, which is caused by excessive occlusion of feature parts, leading to a lack of sufficient features. Our final approach involved using an $\alpha$ value of 0.2. Unlike the fusion part which eliminates interference as much as possible, the estimation module still needs to retain sufficient feature information, thus a threshold is leveraged to constrain the weights of motion reject masks.

\noindent \textbf{Enhanced Motion Mask}: 
    The mask is added in loss function for a direct guidance for the model to focusing on dominant plane. We conduct an experiment to assess the performance of these masks individually, and the results are presented in Table~\ref{tab:Ablation Studies}. Comparing row 4, row 5 and row 7 with row 3 demonstrates that using masks in loss function of homography estimation can significantly reduce errors by helping the network focus on the dominant plane. When comparing row 4 and row 5 with row 7, the fused mask proves to be more effective than using the masks independently. As Fig.~\ref{fig_masks} shows, the $M_{m}$ has a more clear rejection of outlier regions while the $M_{c}$ is closer to features. The fused $M_{e}$ can merge the advantages of both masks as shown in Fig.~\ref{fig_masks} (d), enabling a more precise focus on the dominant plane, thereby enhancing the efficacy of homography estimation.

\noindent \textbf{NLL loss}: 
    Mask generation is enforced using the negative likelihood loss (NLL) function with MVs and homography. A mask is constructed to be one in regions where MVs and homography share similar values, indicating the dominant plane. In this study, we substituted the NLL loss with triplet loss, a method proposed by CAHomo~\cite{cahomo}. The results are presented in Table ~\ref{tab:Ablation Studies}. Comparison between row 6 and row 7 reveals that mask constraints by the NLL loss are more effective than employing triplet loss, leading to a decrease in error from 0.39 to 0.31. Masks lacking NLL constraints tend to converge to all zeros, resulting in minimized triplet loss. Conversely, masks with NLL constraints accurately focus on the dominant plane, as depicted in Fig.~\ref{fig_nll}. Triple loss identifies the dominant plane by masking various areas within the warped features, resulting in inaccuracies and lack of control. In contrast, NLL can more effectively pinpoint the dominant plane through the construction of a probability model using MVs and homography.

\section{Limitation}

While our method excels in achieving state-of-the-art performance in small baseline scenes compared to existing methods, it has limitations when applied to large baseline scenes, particularly when leveraging MVs obtained from video coding as a prior. Motion search for these vectors is computationally demanding in video encoding, prompting various efforts to enhance efficiency, typically capping the search range at 16 pixels to balance performance and computational load. When motion extends beyond this search range, blocks with the lowest BD-rate within the search area are selected. Consequently, these MVs cannot accurately represent true motion. Using these low-quality MVs as a prior for homography estimation generates an all-zero mask to exclude interrupted information, resulting in no positive contribution to our framework. We will address the solution for large baseline alignment in future work.

\section{Conclusion}

We have presented a novel framework, CodingHomo, for unsupervised homography estimation by combining MVs. Our approach incorporates a mask-guided fusion module to merge MVs and homography, as well as a mask-guided homography estimation block to improve homography accuracy. Furthermore, we generate an enhanced dominant plane mask using MVs with a negative likelihood loss to direct the homography estimator's focus towards the dominant plane. The results demonstrate that our method not only achieves state-of-the-art performance but also exhibits robust generalizability. 


 \begin{IEEEbiography}[{\includegraphics[width=1in,height=1.25in,clip,keepaspectratio]{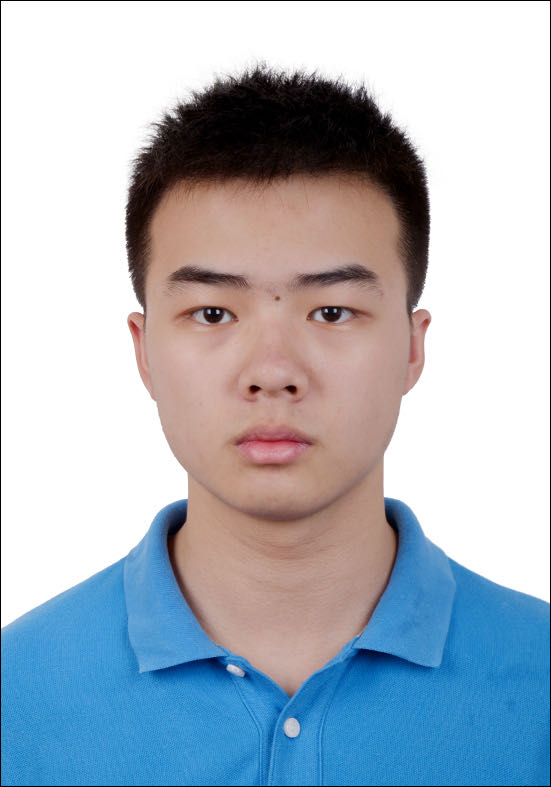}}]{Yike Liu}
     received the B.E. degrees from the University of Electronic Science and Technology of China (UESTC), Chengdu, China, in 2020. Currently, he is a Ph.D. student at School of Information and Communication Engineering, University of Electronic Science and Technology of China, Chengdu, China. His research interests include video processing and computer vision.
  \end{IEEEbiography}

\vspace{-15mm}

\begin{IEEEbiography}
[{\includegraphics[width=1in,height=1.25in,clip,keepaspectratio]{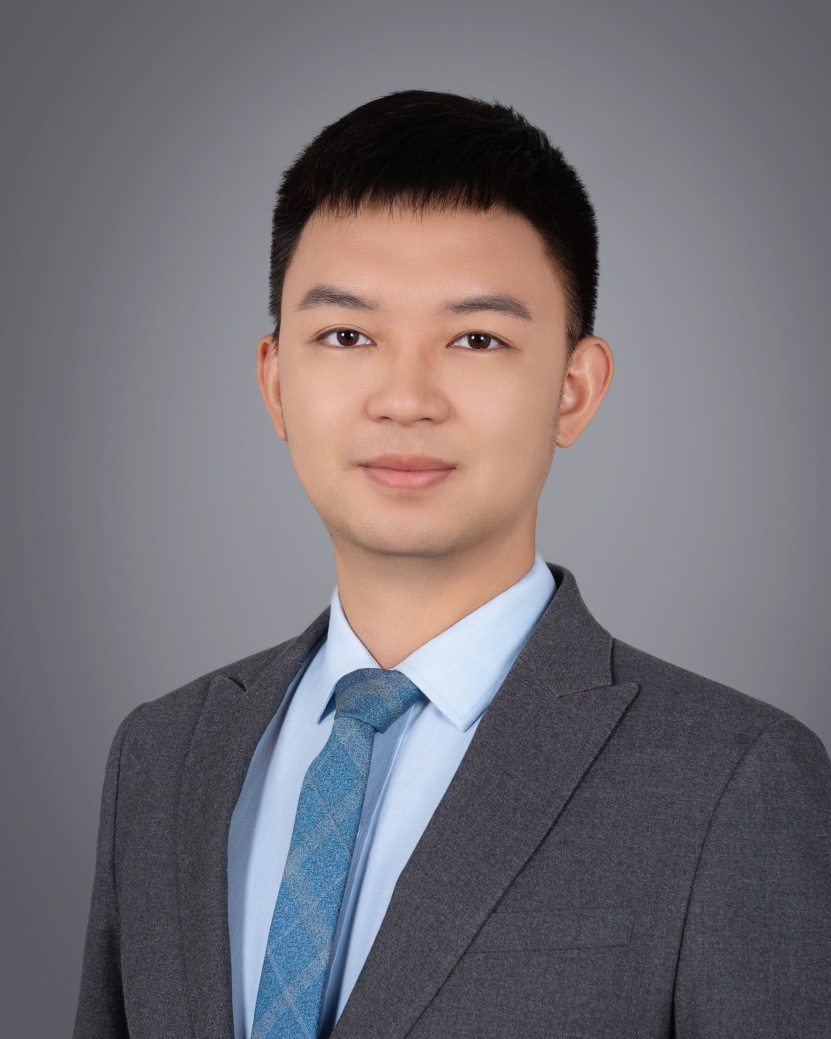}}]{Haipeng Li} received the BEng degree from the University of Electronic Science and Technology of China (UESTC), Chengdu, China, in 2017 and the MSc degree from the Institut Mines-Telecom Atlantique Bretagne Pays de la Loire, Brest, France, in 2020. He was a Researcher in Megvii Research Chengdu during 2019-2022. He has been a PhD student since 2022 in the School of Information and Communication Engineering, UESTC. His research interests include generative models and computer vision. In the past three years, he has published several papers in top journals and conferences such as IJCV, IEEE TIP and TCSVT, ACM ToG, CVPR, ICCV, and AAAI.
\end{IEEEbiography}

\vspace{-15mm}

\begin{IEEEbiography}[{\includegraphics[width=1in,height=1.25in,clip,keepaspectratio]{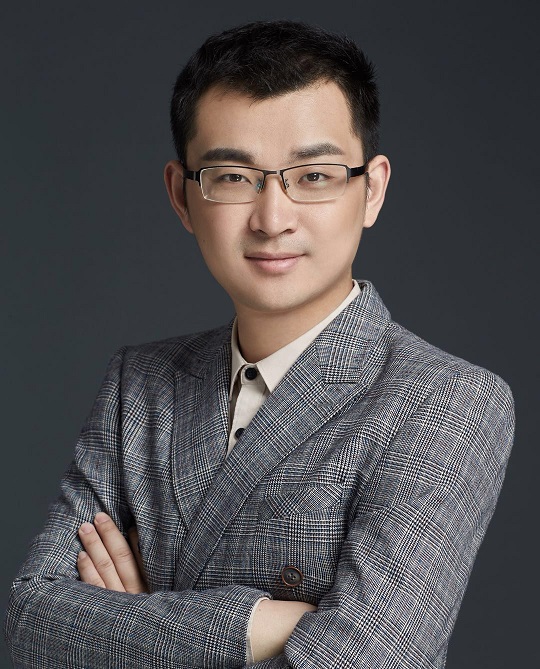}}]{Shuaicheng Liu} (M’14-SM’23) received the Ph.D. and M.Sc. degrees from the National University of Singapore, Singapore, in 2014 and 2010, respectively, and the B.E. degree from Sichuan University, Chengdu, China, in 2008. In 2015, he joined the University of Electronic Science and Technology of China and is currently a Professor with the Institute of Image Processing, School of Information and Communication Engineering, Chengdu, China. His research interests include computer vision and computer graphics.
\end{IEEEbiography}

\vspace{-15mm}

\begin{IEEEbiography}[{\includegraphics[width=1in,height=1.25in,clip,keepaspectratio]{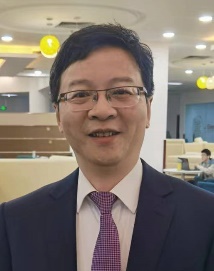}}]{Bing Zeng}
(M’91-SM’13-F’16) received the BEng and MEng degrees in electronic engineering from University of Electronic Science and Technology of China (UESTC), Chengdu, China, in 1983 and 1986, respectively, and the PhD degree in electrical engineering from Tampere University of Technology, Tampere, Finland, in 1991.

He worked as a postdoctoral fellow at University of Toronto from September 1991 to July 1992 and as a Researcher at Concordia University from August 1992 to January 1993. He then joined the Hong Kong University of Science and Technology (HKUST). After 20 years of service at HKUST, he returned to UESTC in the summer of 2013. At UESTC, he leads the Institute of Image Processing to work on image and video processing, multimedia communication, computer vision, and AI technology, was Dean of Glasgow College (a joint school between UESTC and University of Glasgow) during 2018-2022, and has been a Vice Chair of Committee for Academic Affairs since 2016. 
 
He served as an Associate Editor for IEEE TCSVT for 8 years and received the Best Associate Editor Award in 2011. He was General Co-Chair of IEEE VCIP-2016 and PCM-2017. He received a 2nd-Class Natural Science Award (the 1st recipient) from Chinese Ministry of Education in 2014 and was elected as an IEEE Fellow in 2016 for contributions to image and video coding.
\end{IEEEbiography}

\end{document}